\documentclass{article}
\usepackage{spconf,amsmath,graphicx}
\usepackage{multirow}
\usepackage{graphicx}
\usepackage{amssymb}

\title{Unsupervised Extractive Summarization with Heterogeneous Graph
Embeddings for Chinese Document}

\name{Chen Lin \qquad Ye Liu \qquad Siyu An \qquad Di Yin}

\address{Tencent}
      
\begin{document}
%
\maketitle
\begin{abstract}

In the scenario of unsupervised extractive summarization, learning high-quality sentence representations is essential to select salient sentences from the input document. Previous studies focus more on employing statistical approaches or pre-trained language models (PLMs) to extract sentence embeddings, while ignoring the rich information inherent in the heterogeneous types of interaction between words and sentences. In this paper, we are the first to propose an unsupervised extractive summarizaiton method with heterogeneous graph embeddings (HGEs) for Chinese document. A heterogeneous text graph is constructed to capture different granularities of interactions by incorporating  graph structural information. Moreover, our proposed graph is general and flexible where additional nodes such as keywords can be easily integrated. Experimental results demonstrate that our method consistently outperforms the strong baseline in three summarization datasets. 





\end{abstract}
\section{Introduction}
\label{sec:intro}
Text summarization is the task of automatically condensing the input document to a shorter version while maintaining its most important information. The ability to condense text information can be applied to various applications such as online news services, answering questions, and opinion mining. Basically, there are two main types of text summarization tasks: extractive and abstractive. Extractive text summarization \cite{cheng2016neural, nallapati2017summarunner, narayan2018ranking, wang2020heterogeneous, liu2021hetformer, yang2022hierarchical} directly selects some sentences from the input document to assemble summary. In contrast, abstractive text summarization \cite{rush2015neural, durrett2016learning, tu2016modeling, liu2019text, liu2021simcls} can generate novel words and phrases which didn't appear in the input document.

To effectively extract important sentences from the input document, we first need to prepare large-scale datasets which consist of a large number of document-summary pairs. However, it is obviously impractical to obtain large-scale and high-quality datasets for different text summarization tasks as human annotation becomes a bottleneck. Therefore, unsupervised learning approaches have become the main areas of interest among researchers. Previous efforts \cite{mihalcea2004textrank, khan2019extractive} represent the sentences in the input document as the nodes of a graph and use graph-based ranking algorithms \cite{kleinberg1999authoritative, page1999pagerank} to decide the importance of each node in the graph. Although easy to implement and computationally inexpensive, graph-based ranking algorithms usually measure the similarity between nodes based on statistical approaches \cite{harris1954distributional, luhn1957statistical} which did not take into account the semantic meanings of context words.

To better capture sentential meanings, pre-trained language models (PLMs) are integrated into unsupervised extractive summarization methods. PacSum\cite{zheng2019sentence} is one of the most promising approaches.
Although PLMs-based methods have achieved state-of-the-art results on several summarization datasets, few of them have taken into account the various relations between words and sentences of the input document.

As documents are composed of various components such as words, phrases, sentences and paragraphs, it is a challenging task to develop an effective model for text representations that can considerably capture the inherent characteristics of texts \cite{osman2020graph}. To overcome the limitations of previous studies \cite{mihalcea2004textrank, khan2019extractive, zheng2019sentence}, we revisit the problem of how to represent texts in the shape of a graph to better exploit the underlying relationships between different components of the input document. Traditional methods represent texts as homogeneous graphs which only consist of sentence nodes and make use of statistical approaches or PLMs to compute the semantic similarity between sentence nodes. In this paper,  we introduce a heterogeneous text graph to effectively model the relationships and structures of texts. In our proposed heterogeneous text graph, 
different granularities of  sentence nodes and word nodes are utilized to extract hierarchical structural information,  and three types of edges are introduced in the graph. 
As a result, sentence nodes can interact with each other in light of the structural overlap of word nodes. More generally, additional nodes, such as keywords or named entities, can be easily integrated into our proposed graph. We highlight our contributions as follows.

\begin{itemize}
\item To our knowledge, we are the first to construct a heterogeneous text graph to model the heterogeneous types of interaction between words and sentences in the unsupervised extractive summarization task for Chinese documents. To enhance sentence representations, hierarchical rich structural information are incorporated by utilizing different granularities of nodes.

\item Our proposed heterogeneous text graph is general and flexible where additional nodes, such as keywords, can be easily integrated. 

\item We conducted extensive experiments to evaluate the effectiveness of our proposed method. The experimental results indicate that our model is competitive compared to the SOTA algorithms in three datasets.
\end{itemize}

\section{Methodology}
\label{sec:method}

In this section, an overview of  our unsupervised extractive summarization framework is presented. We first introduce how to construct a heterogeneous text graph based on multiple types of word nodes and sentence nodes. After then, we present metapath-based random walk method to train our unsupervised extractive summarization model. Finally, a centrality metric is given to select salient sentences from the input document as the output summary.

\begin{figure}[h]
  \centering
  \includegraphics[width=1\linewidth]{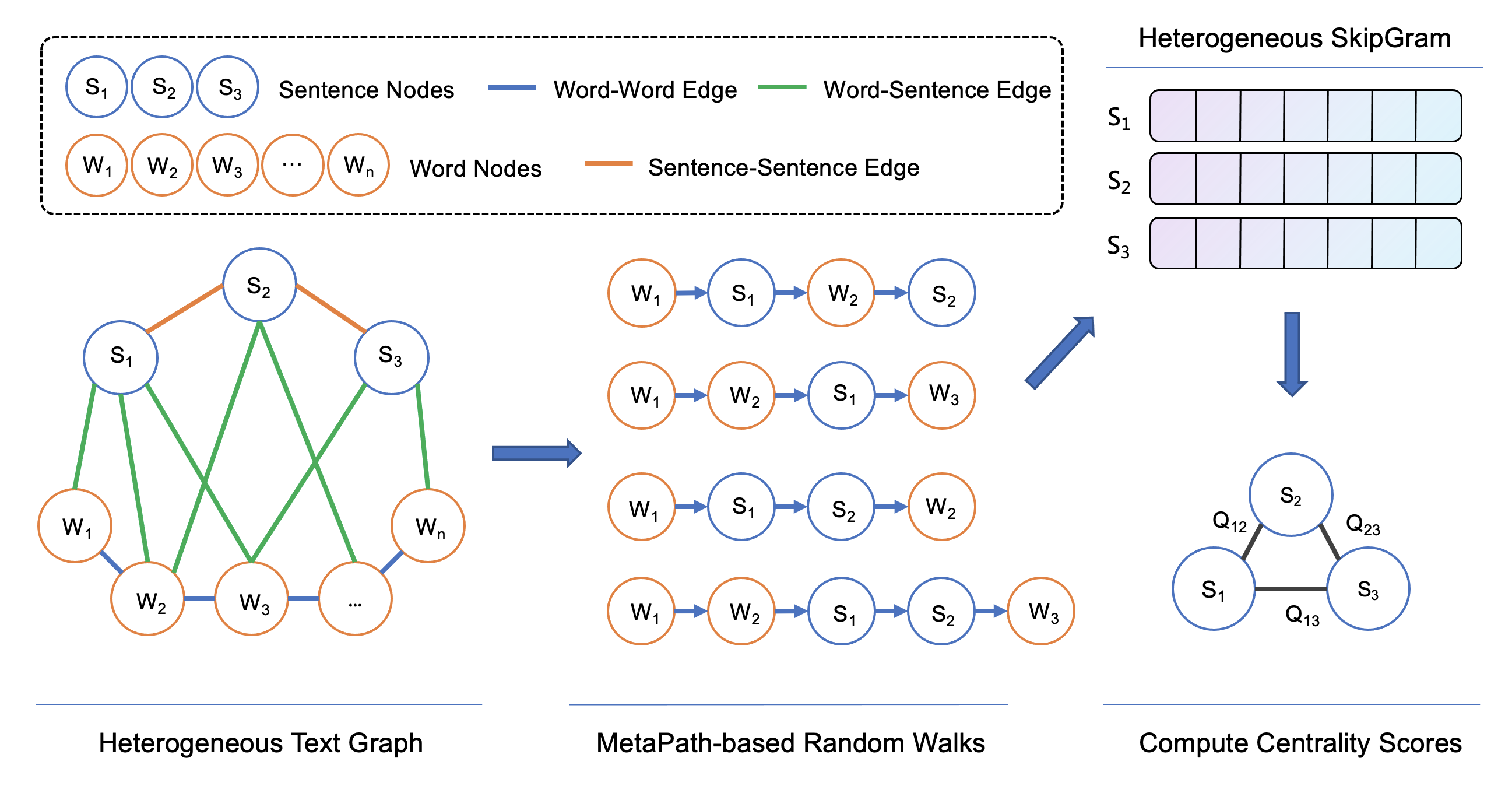}
  \caption{An overview of our proposed unsupervised extractive summarization with HGEs. }
  \label{FIG:ues}
\end{figure}

\subsection{Heterogeneous Text Graph}

Figure \ref{FIG:ues} presents an overview of our proposed unsupervised extractive summarization with heterogeneous graph embeddings for Chinese document. 
As a preprocessing step, we first filter some useless words from the sentences, such as stopwords, punctuations, which cannot add much meaning and may bring the noisy signals to the model. 
Then, we divide the input document into sentences and each sentence is tokenized into a list of individual words.

To capture the interactions among sentences and words, we introduce three types of edges. Given a document $D$, we can represent a document as a graph $G = (V, E)$, where $V$ is the set of nodes and $E$ is the set of edges between the nodes. Our heterogeneous text graph consists of two types of nodes: words and sentences. Therefore, it can be formally defined as $V = V_w \cup V_s$, where $V_w = \{w_1, ..., w_n\}$ denotes $n$ unique words in the input document and $V_s = \{s_1, ..., s_m\}$ denotes $m$ different sentences in the input document. In order to learn the heterogeneous relations, there are three types of edges in the heterogeneous text graph which can be defined as $E= E_{ww} \cup E_{ws} \cup E_{ss}$, where $E_{ww}$ denotes the contextual relations between words, $E_{ws}$ denotes the co-occurence relations between words and sentences, $E_{ss}$ denotes the contextual relations between sentences. 

\textbf{Word-Word Edge (W-W)}\quad We connect the edges between all consecutive words within a sliding window in each sentence. Therefore, we can learn to predict the current word from a window of surrounding context words in a similar way of Word2Vec which is proven to be successful on a variety of downstream NLP tasks. In this way, we can model the semantic relations between words in each sentence.

\textbf{Word-Sentence Edge (W-S)}\quad In order to capture the cross-sentence relationships, we connect each word with the sentence contained in it. Therefore, these word nodes now act as the intermediary between the sentences and enrich the modeling ability of our proposed graph. To be more specific, the word nodes which have more edges than other nodes may have more important meanings in the whole document and the sentences which contain such word nodes is more likely to appear in the output summary.

\textbf{Sentence-Sentence Edge (S-S)}\quad The discourse coherence is very important because it encompasses how sentences are connected, as well as how the entire document is organized to convey information to the readers. It is supposed that the neighbors of current sentence have similar semantic meanings. Therefore, we connect consecutive sentences within a sliding window to explicitly model these relationships.


\subsection{Sentence Embedding Learning}
\subsubsection{Metapath-based Random Walks}
To fully exploit the semantic and structure information of the heterogeneous text graph, we carefully design various metapath schemas, following the process adopted by \cite{fu2020magnn}.
A metapath
schema is defined as a path denoted in the following form.
\begin{equation}
P = N_{1} \overset{R_{1} }{\rightarrow} N_{2} \overset{R_{2} }{\rightarrow} ... \overset{R_{l} }{\rightarrow} N_{l+1}
\end{equation}
where $R = R_{1} \circ R_{2} \circ ... \circ R_{l}$ defines the composite relations between node types $V_{1}$ and $V_{l}$ \cite{sun2012mining}. After defining the schema of metapath, we can generate multiple sequences of nodes originating from the same starting node, following the same metapath schema. We can call each such sequence a metapath instance of $P$. 


\subsubsection{Heterogeneous SkipGram}
Given the generated instances of metapath schemas, our goal is to maximize the likelihood of preserving both the structures and semantics of a given heterogeneous graph. Therefore, we use metapath-based random walks to simultaneously learn the low-dimensional and latent embeddings for multiple types of nodes. Then we use the heterogeneous skip-gram model to learn effective node representations for the heterogeneous text graph $G = (V, E)$ by maximizing the probability of having the heterogeneous context $N_t(v)$, $t \in T_V$ given a node $v$:
\begin{equation}
arg\max_{\theta} \sum_{v \in V}^{} \sum_{t \in T_{V} }^{} \sum_{c_t \in N_t(v)}^{} \log p(c_t | v;\theta)
\end{equation}
where $N_t(v)$ denotes $v$ 's neighborhood with the $t$-th type of nodes and p($c_{t}$$|$$v$;$\theta$) is adjusted to the specific node type t, that is,
\begin{equation}
p(c_t|v;\theta)=\frac{e^{X_{c_t} \cdot X_v}}{ {\textstyle \sum_{u_t \in V_t}^{} e^{X_{u_t} \cdot X_v} } } 
\end{equation}
where $X_v$ is the $v$-th row of $X$, representing the embedding vector for node $v$.
By further leveraging the heterogeneous negative sampling technique, we can achieve the following objective.
\begin{equation}
O(x)=\log_{}{\sigma(X_{c_t} \cdot X_v) + \sum_{m=1}^{M} E_{u_t^m \sim P_t(u_t) [\log \sigma(-X_{u_t^m}) \cdot X_v]}}
\end{equation}

\subsection{Centrality-based Summarization}

Finally, we need to select salient sentences from the input document to assemble summaries. To take advantage of both graph structural information and semantic information, we concat the graph representation and the sentence semantic representation as $v_{i}$. We employ a pairwise dot product to compute an unnormalized similarity matrix $\overline{Q}$.
The normalized similarity matrix Q is defined based on $\overline{Q}$:

\begin{equation}
\label{equation:overline-Q}
\overline{Q}_{ij} = \overline{Q}_{ij} - \left [ \min \overline{Q} + \beta (\max \overline{Q} - \min \overline{Q})\right ]
\end{equation}

\begin{equation}
Q_{ij} =\left\{
\begin{aligned}
\overline{Q}_{ij} & , & if \quad \overline{Q}_{ij} > 0 \\
0 & , & otherwise
\end{aligned}
\right.
\end{equation}
Due to the fact that all possible sentence pairs may be assigned high values in some cases, we use Equation (\ref{equation:overline-Q}) to emphasize the relative contribution of different similarity scores. This is particularly important when computing the similarity matrix from a practical point of view. 

The importance score of each sentence can be computed using the centrality based on the directed graph which consists of only sentence nodes as follows:
\begin{equation}
\label{equation:centrality}
\lambda_{1} \sum_{j<i}^{} Q_{ij} + \lambda _{2} \sum_{j > i} Q_{ij}
\end{equation}

where $\lambda_{1}$, $\lambda_{2}$ are different weights for the edges directed in forward and backward orientation. 



\begin{table*}[]
\tiny
\begin{center}
\scalebox{0.99}{
\resizebox{\textwidth}{!}{%
\begin{tabular}{|l|ccc|ccc|ccc|}
\hline
\multirow{2}{*}{\textbf{Model}} &
  \multicolumn{3}{c|}{\textbf{TTNews}} &
  \multicolumn{3}{c|}{\textbf{CNewSum}} &
  \multicolumn{3}{c|}{\textbf{Education}} \\ \cline{2-10} 
 &
  \multicolumn{1}{c|}{R-1} &
  \multicolumn{1}{c|}{R-2} &
  R-L &
  \multicolumn{1}{c|}{R-1} &
  \multicolumn{1}{c|}{R-2} &
  R-L &
  \multicolumn{1}{c|}{R-1} &
  \multicolumn{1}{c|}{R-2} &
  R-L \\ \hline
ORACLE \cite{nallapati2017summarunner} &
  \multicolumn{1}{c|}{45.6} &
  \multicolumn{1}{c|}{31.2} &
  41.7 &
  \multicolumn{1}{c|}{46.8} &
  \multicolumn{1}{c|}{30.5} &
  40.0 &
  \multicolumn{1}{c|}{48.7} &
  \multicolumn{1}{c|}{37.2} &
  43.5 \\ \hline
LEAD &
  \multicolumn{1}{c|}{30.8} &
  \multicolumn{1}{c|}{18.4} &
  24.9 &
  \multicolumn{1}{c|}{30.4} &
  \multicolumn{1}{c|}{17.3} &
  25.3 &
  \multicolumn{1}{c|}{36.1} &
  \multicolumn{1}{c|}{28.6} &
  33.8 \\ \hline
Pointer-Generator \cite{zheng2019sentence} $\dag$ $\star$&
  \multicolumn{1}{c|}{42.7} &
  \multicolumn{1}{c|}{27.5} &
  36.2 &
  \multicolumn{1}{c|}{25.7} &
  \multicolumn{1}{c|}{11.1} &
  19.6 &
  \multicolumn{1}{c|}{39.9} &
  \multicolumn{1}{c|}{31.5} &
  29.7 \\ \hline
TextRank-TFIDF \cite{zheng2019sentence} $\dag$ $\star$&
  \multicolumn{1}{c|}{25.6} &
  \multicolumn{1}{c|}{13.1} &
  19.7 &
  \multicolumn{1}{c|}{24.0} &
  \multicolumn{1}{c|}{13.7} &
  20.1 &
  \multicolumn{1}{c|}{22.5} &
  \multicolumn{1}{c|}{15.8} &
  18.6 \\ \hline
PacSum-BERT \cite{zheng2019sentence} $\dag$&
  \multicolumn{1}{c|}{32.8} &
  \multicolumn{1}{c|}{18.9} &
  26.1&
  \multicolumn{1}{c|}{-} &
  \multicolumn{1}{c|}{-} &
  - &
  \multicolumn{1}{c|}{-} &
  \multicolumn{1}{c|}{-} &
  -
 \\ \hline
PacSum-BERT \cite{zheng2019sentence} $\star$ &
  \multicolumn{1}{c|}{35.2} &
  \multicolumn{1}{c|}{21.0} &
  28.5 &
  \multicolumn{1}{c|}{33.8} &
  \multicolumn{1}{c|}{18.9} &
  27.1 &
  \multicolumn{1}{c|}{32.6} &
  \multicolumn{1}{c|}{23.9} &
  29.9 \\ \hline
Ours (HGEs) &
  \multicolumn{1}{c|}{36.3} &
  \multicolumn{1}{c|}{22.1} &
  30.0 &
  \multicolumn{1}{c|}{34.0} &
  \multicolumn{1}{c|}{19.2} &
  27.3 &
  \multicolumn{1}{c|}{34.8} &
  \multicolumn{1}{c|}{26.3} &
  32.5 \\ \hline
Ours (HGEs + PacSum-BERT) &
  \multicolumn{1}{c|}{\textbf{36.8}} &
  \multicolumn{1}{c|}{\textbf{22.2}} &
  \textbf{30.2} &
  \multicolumn{1}{c|}{\textbf{34.4}} &
  \multicolumn{1}{c|}{\textbf{19.6}} &
  \textbf{28.0} &
  \multicolumn{1}{c|}{\textbf{36.2}} &
  \multicolumn{1}{c|}{\textbf{28.5}} &
  \textbf{34.0} \\ \hline
\end{tabular}%
}
}
\caption{Results on the TTNews, CNewSum and Education. $\dag$: results come from \cite{zheng2019sentence}; $\star$: results come from our re-implementation; $\dag$ $\star$: the results of TTNews come from \cite{zheng2019sentence},  others come from our re-implementation.}
\label{tab:results}
\end{center}
\end{table*}

\section{Experiments}

\subsection{DataSets and Metrics}
We performed experiments on three recently released single-document summarization datasets. 
\textbf{TTNews} \footnote{http://tcci.ccf.org.cn/conference/2017/taskdata.php} is created for the shared summarization task at NLPCC 2017. It contains a large set of news articles browsed on Toutiao.com. The news articles come from a large number of different sources and meanwhile contain different topics. 
\textbf{CNewSum} \footnote{https://dqwang122.github.io/projects/CNewSum/} is a large-scale Chinese news summarization dataset collected from hundreds of thousands of news publishers and a team of expert editors is hired to provide human-written summaries for the daily news feed. \textbf{Education} \footnote{https://github.com/wonderfulsuccess/chinese\_abstractive\_corpus} is collected from historical articles in vertical mainstream media in the education and training industry.

We evaluated the quality of the summarization using ROUGE F1\cite{hu2015lcsts}. We report unigram and bigram overlap ROUGE-1 (R-1) and ROUGE-2 (R-2) as means of assessing informativeness and the longest common subsequence ROUGE-L (R-L) as means of assessing fluency.

\begin{table}[]
\scalebox{0.48}{
\tiny
\resizebox{\textwidth}{!}{%
\begin{tabular}{|c|ccc|c|c|c|}
\hline
\multirow{2}{*}{\textbf{Dataset}} & \multicolumn{3}{c|}{\textbf{Edge Type}} & \multirow{2}{*}{\textbf{R-1}} & \multirow{2}{*}{\textbf{R-2}} & \multirow{2}{*}{\textbf{R-3}} \\ \cline{2-4}
                                    & \multicolumn{1}{c|}{W-W} & \multicolumn{1}{c|}{W-S} & S-S &      &      &      \\ \hline
\multirow{4}{*}{TTNews}    & \multicolumn{1}{c|}{}             & \multicolumn{1}{c|}{\checkmark}             &              & 36.1 & 21.9 & 29.8 \\ \cline{2-7} 
                                    & \multicolumn{1}{c|}{\checkmark}             & \multicolumn{1}{c|}{\checkmark}             &              & 36.2 & 21.7 & 29.7 \\ \cline{2-7} 
                                    & \multicolumn{1}{c|}{}             & \multicolumn{1}{c|}{\checkmark}             &     \checkmark         & 36.6 & 22.1 & 30.1 \\ \cline{2-7} 
                                    & \multicolumn{1}{c|}{\checkmark}             & \multicolumn{1}{c|}{\checkmark}             &      \checkmark        & \textbf{36.8} & \textbf{22.2} & \textbf{30.2} \\ \hline
\multirow{4}{*}{CNewSum}   & \multicolumn{1}{c|}{}             & \multicolumn{1}{c|}{\checkmark}             &              & 34.0 & 19.2 & 27.6 \\ \cline{2-7} 
                                    & \multicolumn{1}{c|}{\checkmark}             & \multicolumn{1}{c|}{\checkmark}             &              & 34.3 & 19.3 & 27.8 \\ \cline{2-7} 
                                    & \multicolumn{1}{c|}{}             & \multicolumn{1}{c|}{\checkmark}             &     \checkmark         & 34.2 & 19.3 & 27.9 \\ \cline{2-7} 
                                    & \multicolumn{1}{c|}{\checkmark}             & \multicolumn{1}{c|}{\checkmark}             &       \checkmark       & \textbf{34.4} & \textbf{19.6} & \textbf{28.0} \\ \hline
\multirow{4}{*}{Education} & \multicolumn{1}{c|}{}             & \multicolumn{1}{c|}{\checkmark}             &              & 32.5 & 25.0 & 29.8 \\ \cline{2-7} 
                                    & \multicolumn{1}{c|}{\checkmark}             & \multicolumn{1}{c|}{\checkmark}             &              & 33.5 & 24.6 & 30.4 \\ \cline{2-7} 
                                    & \multicolumn{1}{c|}{}             & \multicolumn{1}{c|}{\checkmark}             &     \checkmark         & 33.4 & 24.6 & 31.8 \\ \cline{2-7} 
                                    & \multicolumn{1}{c|}{\checkmark}             & \multicolumn{1}{c|}{\checkmark}             &     \checkmark         & \textbf{36.2} & \textbf{28.5} & \textbf{34.0} \\ \hline
\end{tabular}%
}
}
\caption{Ablation studies on types of edge.}
\label{tab:edge-types}
\end{table}



\subsection{Baselines}
We compare our proposed approach with previous unsupervised models in extractive summarization. In addition, we report the LEAD baseline which directly selects the first k sentences(k = 1)  from the input document as the output summary and neural abstractive approaches for completeness. For extractive summarization approaches, we rank all sentences using Equation(\ref{equation:centrality}) and select the top k sentence (k = 1) as the summary for fair comparison when generating the summary for a new document during the test time.

TextRank\cite{mihalcea2004textrank} is one of the most classic unsupervised extractive summarization methods. The edges between sentence nodes in the graph are defined by the similarities between nodes based on tf-idf. Subsequently, it selects sentences based on a simple variant of the PageRank \cite{page1999pagerank} algorithm. PacSum \cite{zheng2019sentence} is another unsupervised extractive summarization algorithm that uses the BERT model as an encoder for all sentences. Sentences are ranked using centrality based on a sentence graph. Pointer-generator \cite{see2017get} is a supervised abstractive summarization method which can copy words from the source text while retaining the ability to produce novel words. At the same time, we refer to \cite{nallapati2017summarunner} to obtain Oracle results using a greedy algorithm.

\subsection{Results}
In this section, we present the results of our model compared to baselines with respect to ROUGE metrics. As shown in Table \ref{tab:results}, we present the results of the \textbf{ORACLE} and \textbf{LEAD} in top two lines. We can tell from the results that the LEAD method achieved a comparable result although it is a simple method due to lead bias is a common phenomenon in extractive summarization. The third line in the table presents the results of \textbf{Pointer-Generator} which is a supervised abstractive model for a complete comparison. The 4-8th lines in the table present the results of unsupervised extractive summarization, including the methods \textbf{TextRank} and \textbf{PacSum}. We reimplemented the \textbf{PacSum} method based on HuggingFace framework. As the table shows, the \textbf{PacSum} achieves a significant better results than \textbf{TextRank}. The last two lines of the table present the results of \textbf{Our} approaches which use words and sentences to construct the heterogeneous text graph. The experimental results show that our method that incorporates Bert embeddings and HGEs outperforms the PacSum method.

\begin{table}[]
\scalebox{0.48}{
\tiny
\resizebox{\textwidth}{!}{%
\begin{tabular}{|c|ccc|c|c|c|}
\hline
\multirow{2}{*}{\textbf{Dataset}} & \multicolumn{2}{c|}{\textbf{Node Type}} & \multirow{2}{*}{\textbf{R-1}} & \multirow{2}{*}{\textbf{R-2}} & \multirow{2}{*}{\textbf{R-3}} \\ \cline{2-3}
                                    & \multicolumn{1}{c|}{Word} & \multicolumn{1}{c|}{Keyword}  &      &      &      \\ \hline
\multirow{2}{*}{TTNews}    & \multicolumn{1}{c|}{\checkmark}              & \multicolumn{1}{c|}{}                            & 36.8 & 22.2 & 30.2 \\ \cline{2-6} 
                                    & \multicolumn{1}{c|}{}              & \multicolumn{1}{c|}{\checkmark}                       & \textbf{37.0} & \textbf{22.3} & \textbf{30.4} \\ \hline
\multirow{2}{*}{CNewSum}   & \multicolumn{1}{c|}{\checkmark}              & \multicolumn{1}{c|}{}                            & 34.4 & 19.6 & 28.0 \\ \cline{2-6} 
                                    & \multicolumn{1}{c|}{}              & \multicolumn{1}{c|}{\checkmark}                     & \textbf{34.5} & \textbf{19.7} & \textbf{28.2} \\ \hline
\multirow{2}{*}{Education} & \multicolumn{1}{c|}{\checkmark}              & \multicolumn{1}{c|}{}                           & 36.2 & 28.5 & 34.0 \\ \cline{2-6} 
                                    & \multicolumn{1}{c|}{}              & \multicolumn{1}{c|}{\checkmark}                       & \textbf{36.5} & \textbf{28.8} & \textbf{34.3} \\ \hline
\end{tabular}%
}
}
\caption{Ablation studies on types of node.}
\label{tab:node-types}
\end{table}

\subsection{Ablation Study}
\quad \quad \textbf{Edge Types}\quad Table \ref{tab:edge-types} presents the ablation study to assess the relative contributions of the different types of edges in the heterogeneous text graph. We keep all the hyperparameters unchanged with respect to the best settings and only vary the types of edge. We can see that adding Word-Word edges and Sentence-Sentence edges to the heterogenous text graph boosts the model's performance when comparing with the graph with Word-Sentence edges only. Moreover, we find that the improvement of combining various types of edge are stable across multiple datasets.

\textbf{Node Types}\quad Table \ref{tab:node-types} reports the ablation results with different word nodes. Our proposed graph is general and flexible, where additional nodes can be easily integrated. Due to the major idea of the input document can be highlighted by keywords, we can use keywords as the indicators for salient sentences selection. We utilize Jieba \footnote{https://github.com/fxsjy/jieba} to extract top 20 keywords from documents. Instead of constructing a heterogeneous text graph with word nodes and sentence nodes, we can first extract keywords and then construct a heterogeneous text graph based on these keyword nodes and sentence nodes. As shown in Table \ref{tab:node-types}, we can see that the variant of a heterogeneous text graph with keyword nodes and sentence nodes achieves a better performance.

\section{Conclusion}

In this paper, we propose an unsupervised extractive summarization method with heterogeneous graph embeddings for Chinese document. The introduction of various types of nodes and edges in the graph helps our model to build more complex relationships between sentences. Furthermore, our models have achieved the best results on three datasets compared with the SOTA method. In the future, we will investigate whether the ideas introduced in this paper are applicable to English and multi-document summarization.

\vfill\pagebreak

\bibliographystyle{IEEEbib}
\bibliography{icassp}

\end{document}